\documentclass[12pt,a4paper,twoside]{article}

\usepackage{times}
\usepackage{verbatim}
\usepackage[latin1]{inputenc}
\usepackage[T1]{fontenc}
\usepackage{graphicx}

\usepackage{taln2006}
\usepackage[frenchb]{babel}

\renewcommand\sectionmark[1]{}
\renewcommand\subsectionmark[1]{}

\begin{document}

\title{Outilex, plate-forme logicielle de traitement de textes écrits}

\author{Olivier Blanc, Matthieu Constant, {\'{E}}ric Laporte\\
  IGM - Université de Marne-la-Vallée \\
5, bd Descartes - 77454 Marne-la-Vallée CEDEX 2\\
  olivier.blanc@univ-mlv.fr,
matthieu.constant@univ-mlv.fr,
eric.laporte@univ-mlv.fr\\
}

\date{}

\maketitle

\resume{La plate-forme logicielle Outilex, qui sera mise à la disposition de la
recherche, du développement et de l'industrie, comporte des composants logiciels
qui effectuent toutes les opérations fondamentales du
traitement automatique du texte écrit : traitements sans lexiques, exploitation
de lexiques et de grammaires, gestion de ressources linguistiques. Les données
manipulées sont structurées dans des formats XML, et
également dans d'autres formats plus compacts, soit lisibles soit binaires,
lorsque cela est nécessaire ; les convertisseurs de formats nécessaires sont
inclus dans la plate-forme ; les formats de grammaires permettent
de combiner des méthodes statistiques avec des méthodes fondées sur des
ressources linguistiques. Enfin, des lexiques du français et de l'anglais issus
du LADL, construits manuellement et d'une couverture
substantielle seront distribuées avec la plate-forme sous licence LGPL-LR.
}

\abstract{The Outilex software platform, which will be made available to
research, development and industry, comprises software components implementing
all the fundamental operations of written text processing: processing without
lexicons, exploitation of lexicons and grammars, language resource
management. All data are structured in XML formats, and also in more compact
formats, either readable or binary, whenever necessary; the required format
converters are included in the platform; the grammar formats allow for combining
statistical approaches with resource-based approaches. Manually constructed
lexicons for French and English, originating from the LADL, and of substantial
coverage, will be distributed with the platform under LGPL-LR license.
}

\motsClefs{Analyse syntaxique, motifs lexico-syntaxiques, analyse lexicale,
ressources linguistiques, formats d'échange, automates finis, réseaux de
transitions récursifs}
{Syntactic parsing, lexico-syntactic patterns, lexical analysis, language
resources,
exchange formats, finite-state automata, recursive transition networks}

%% Aller à la page suivante si nécessaire
\newpage
%%================================================================
\section{Introduction}
Le projet Outilex\footnote{Ce travail a été en partie financé par le Ministère
de l'Industrie et par le CNRS.}
vise à mettre à la disposition de la recherche, du
développement et de l'industrie une plate-forme logicielle 
de traitement des langues naturelles ouverte et compatible avec l'utilisation
d'XML, d'automates finis et de ressources linguistiques. En
raison de son ambition internationale, Outilex a également
participé aux efforts actuels de définition de normes en matière de modèles de
ressources linguistiques. Nous présentons ses principaux résultats.

Les méthodes
de traitement des langues naturelles sont encore aujourd'hui, la plupart du
temps, mises en oeuvre par des logiciels dont la diffusion est limitée. De plus,
on dispose rarement de formats d'échange ou de convertisseurs de formats qui
permettraient de combiner plusieurs composants logiciels pour un même
traitement. Quelques plates-formes font exception à cette situation générale,
mais aucune n'est satisfaisante. Intex \cite{Silberztein:1993}, FSM \cite{Mohri:98}
 et Xelda\footnote{http://www.dcs.shef.ac.uk/~hamish/dalr/baslow/xelda.pdf.}
sont fermés au développement collaboratif.
Unitex \cite{Paumier:2003}, inspiré d'Intex mais dont le code source est pour la
quasi-totalité sous licence LGPL\footnote{Lesser General Public License,
http{:}//www.gnu.org/copyleft/lesser.html.},
ne fournit pas de formats XML. Les systèmes
NLTK \cite{LoperBird:2002} et Gate
\cite{Cunningham:2002} n'ont
pas de fonctionnalités de gestion de ressources lexicales.

Le projet Outilex regroupe 10 partenaires français, dont 4 académiques et 6
industriels\footnote{Outre le développement de la plate-forme libre, placée
pour l'essentiel sous la responsabilité des partenaires IGM et Systran, le
projet comporte la réalisation de démonstrateurs propriétaires, dont nous ne
parlerons pas ici.}. Il est coordonné par l'IGM et financé par le ministère de
l'Industrie dans le cadre du Réseau national des technologies logicielles
(RNTL). Préparé sous la direction de Maurice Gross, il a été lancé en 2002 et
doit se terminer en 2006. 

Les modules d'Outilex sont intégrés sous la forme d'une interface graphique
programmée en Java qui appelle des programmes en C++. L'interface permet de
travailler sur différents projets regroupant un ensemble
de ressources (textes, dictionnaires et grammaires).
L'utilisateur définit une chaîne de traitement. Les entrées,
les sorties et les résultats intermédiaires sont visualisables dans un
conteneur à onglets dédié à cet effet.
\section{Traitement sans lexiques}
Nous regroupons dans cette partie les opérations qui ne font pas appel à des
informations extraites de ressources lexicales.
Ces opérations ont pour résultat une représentation du texte comme
séquence de tokens\footnote{Nous n'avons pas inclus dans la plate-forme de traitements
applicatifs opérant sur ce modèle de texte, ni sur le modèle encore plus simple du sac de
tokens, mais de tels traitements nous semblent facile à interfacer, en raison
justement de la simplicité des modèles sous-jacents.}.
Le module qui les met en oeuvre prend en entrée un texte brut
ou HTML et il produit en sortie le texte segmenté en paragraphes, en
phrases et en tokens dans un format XML (seg.xml) proche de celui proposé par
le projet de norme ISO d'annotation morpho-syntaxique de textes (MAF)
\cite{ClementClergerie:LTC2005} élaboré dans le cadre du projet
RNIL\footnote{Ressources normalisées en ingénierie des langues.}.
Les éventuelles balises de mise en page HTML sont
conservées et placées dans des éléments XML qui les distinguent
des données textuelles.
\begin{figure}[h]
\footnotesize
<?xml version="1.0"?>
<document original\_format="txt"><par id="1"><tu id="s0"><token type="word"
id="t1" alph="latin" case="capit">La</token> <token type="word" id="t2"
alph="latin">police</token> <token type="word" id="t3" alph="latin">a</token>
<token type="word" id="t4" alph="latin">saisi</token> <token type="numeric"
id="t5">164</token> ... <token type="punctuation"
id="t11">.</token></tu></par>
</document>
\caption{Texte segmenté au format seg.xml}
\label{fig:tokenization}
\end{figure}
Les règles de segmentation en tokens et en phrases sont basées sur la
catégorisation des caractères définie par la norme Unicode (ex. lettres, chiffres).
{\`{A}} chaque token est associé un certain nombre d'informations
telles que son type (mot, nombre, ponctuation, etc.), son
alphabet (latin, grec), sa casse (mot en minuscule, commençant par une
majuscule, etc.) ainsi que d'autres informations pour les autres symboles
(signe de ponctuation ouvrant ou fermant, etc.).
De plus, un identifiant est associé à chaque token qui sera conservé durant toutes
les phases du traitement.
Par exemple, la phrase {\it La police a saisi 164 procès-verbaux jeudi dernier}
est segmentée comme dans la figure~\ref{fig:tokenization}.
Appliqué à un corpus de dépêches AFP (352 464 tokens), ce module traite 22 185 mots
par seconde\footnote{Ce test et les tests suivants
ont été effectués sur un ordinateur de bureau équipé d'un processeur Intel Pentium
cadencé
à 2.8 GHz et de 512 Mo de mémoire RAM.}.
\section{Traitement par lexiques}
\label{dicos}
Les traitements évoqués dans la partie précédente ont pour résultat une
représentation du texte comme séquence de tokens. Nous pensons 
qu'une plate-forme généraliste doit
intégrer certaines notions fondamentales absentes de ce modèle, comme celle de
mots composés ou expressions multi-mots, ou la séparation des emplois en cas
d'ambiguïté. Les produits de la linguistique de corpus seuls \cite{treetagger}
ne sont pas de nature à résoudre les problèmes posés par de telles notions.
L'un des moyens pour y parvenir est l'utilisation de lexiques et de grammaires.
L'utilisation de lexiques par les entreprises du domaine s'est d'ailleurs
largement étendue au cours des dernières années. C'est pourquoi Outilex fournit
un jeu complet de composants logiciels pour les opérations sur les lexiques. De
plus, dans le cadre de sa contribution à Outilex, l'IGM a rendu
publique \footnote{http://infolingu.univ-mlv.fr, suivre Données linguistiques puis
Dictionnaires.}
une proportion substantielle des lexiques
du LADL\footnote{Laboratoire
d'automatique documentaire et linguistique, Université Paris 7, 1968-2000.}
pour le français (109 912 lemmes simples et 86 337 lemmes composés)
\footnote{Le jeu d'étiquettes pour le français combine 13 catégories
morpho-syntaxiques, 18 traits flexionnels et divers traits syntaxico-sémantiques.}
et l'anglais (166 150 lemmes simples et 13 361 lemmes composés).
Ces ressources sont proposées sous la
licence LGPL-LR\footnote{Lesser General Public License for Language Resources,
http://infolingu.univ-mlv.fr/lgpllr.html. Les droits
et devoirs donnés aux utilisateurs par la licence LGPL-LR sont l'équivalent,
pour les ressources linguistiques, de ceux donnés aux utilisateurs de la licence
LGPL pour les logiciels.},
créée dans le cadre d'Outilex et agréée par la FSF\footnote{Free Software
Foundation, http://www.fsf.org/ .}.
Les programmes d'Outilex sont compatibles avec toutes les langues européennes
à flexion par suffixes. Des extensions seront nécessaires pour les autres
types de langues.
\subsection{Formats de lexiques}
L'absence de formats génériques et de documentation sur les données sont deux
obstacles à l'utilisation et à l'échange de lexiques pour le traitement
automatique des langues. Les formats adoptés par Outilex tirent parti de deux
circonstances : d'une part, l'émergence actuelle de modèles de données
consensuels dans les projets de normalisation ;
d'autre part, le fait que l'IGM dispose de lexiques du
français construits manuellement et d'une grande couverture\footnote{Ces deux
circonstances sont liées à des travaux effectués au LADL sous la direction de
Maurice Gross, la première par l'intermédiaire du projet Genelex de
normalisation de lexiques \cite{NormierNossin:1990}, la deuxième à travers le
système de lexiques Dela \cite{Courtois:1990}, \cite{Courtois:2004}.}.
Nous avons traduit en XML le
format Dela et inséré dans les balises de la documentation sur les données. Le
format obtenu pour les lexiques de formes fléchies, dic.xml, illustré par la
fig.~2, est adapté à l'échange de données et compatible avec
le modèle LMF \cite{Francopoulo:2003}.
Le format Dela, plus compact : \textit{appelés du contingent,appelé du
contingent.N+hum:mp}, est adapté à la visualisation sur écran et à la
maintenance manuelle par les linguistes \cite{Laporte:LTC2005}. Nous avons donc
réalisé des convertisseurs dans les deux sens entre ces deux formats. Il existe
en outre un format opérationnel que nous décrirons dans la section~5.2.

\begin{minipage}[b]{2.25in}
\tiny
\begin{verbatim}
<entry>
  <lemma>appelé du contingent</lemma>
  <pos name='noun'/>
  <feat name='subcat' value='human'/>
  <inflected>
    <form>appelés du contingent</form>
    <feat name='gender' value='masculine'/>
    <feat name='number' value='plural'/>
  </inflected>
</entry>
\end{verbatim}
\normalsize
\center{\textsc{Fig.~2} -- Un extrait de lexique au format dic.xml}
\end{minipage}
\ 
\begin{minipage}[b]{4in}
\tiny
\begin{verbatim}
<attrtype name='antepos' type='bool'>
  <true alias='g'/>
</attrtype>
<pos name='adj' cutename='A'>
  <attribute name='postpos' type='postpos' default='false' shortcut='yes'/>
  <attribute name='antepos' type='antepos' default='false' shortcut='yes'/>
  <attribute name='gender' type='gender' shortcut='yes'/>
  <attribute name='number' type='number' shortcut='yes'/>
</pos>
\end{verbatim}
\normalsize
\center{\textsc{Fig.~3} -- Extrait de la description d'un
jeu d'étiquettes au format LingDef pour le français}
\end{minipage}

En pratique, l'utilisation de lexiques permet de manipuler des jeux d'étiquettes
complexes et d'une granularité fine, qui nécessitent une description formelle telle
que celle de la fig.~3.
\setcounter{figure}{3}
\subsection{Consultation des lexiques}
Notre étiqueteur morpho-syntaxique\footnote{Les modules de la chaîne de
traitement ayant été implémentés de manière à être indépendants les uns des
autres, il est possible de substituer un autre étiqueteur à celui proposé par
défaut.}
prend un texte segmenté au format \textit{seg.xml} en
entrée et attribue à chaque forme (simple ou composée) l'ensemble
des étiquettes lui correspondant extraites des lexiques indexés (cf. 5.2).
Il est possible d'appliquer un ensemble de lexiques à un texte dans la
même passe de traitement.
De plus, un système de priorités permet de bloquer des analyses issues de
lexiques à faible priorité si la forme considérée est également présente dans
un lexique de priorité supérieure.  Ainsi, nous fournissons par défaut un
lexique général proposant un grand nombre d'analyses pour la langue standard,
que l'utilisateur peut, pour une application spécifique, enrichir à l'aide de
lexiques complémentaires et/ou filtrer avec un lexique prioritaire.
Enfin, plusieurs options, qui peuvent se combiner entre elles, permettent de
paramétrer la méthode de consultation : il est possible d'ignorer
complètement la casse, ou les accents et autres signes diacritiques.
Ces paramètres permettent d'adapter notre étiqueteur au
texte analysé (article de journal, page internet, e-mail, etc.).
Appliqué au corpus de dépêches AFP (cf. 2) avec les dictionnaires décrits en 3,
Outilex étiquette une moyenne de 6 650 mots par seconde\footnote{4,7 \%
des occurrences de tokens n'ont pas été trouvées dans le dictionnaire~;
cette valeur tombe à 0,4 \% si on déduit le nombre de celles qui commencent
par une majuscule.}.
\subsection{Représentation du texte étiqueté}
L'utilisation exclusive de lexiques pour étiqueter les textes produit des
ambiguïtés lexicales. Le modèle le plus
adapté pour représenter le texte étiqueté dans ces conditions est l'automate
fini acyclique, en général appelé "treillis" dans ce contexte.
La figure~4 présente une partie de l'automate du texte
obtenu après l'étiquetage de la phrase segmentée présentée dans la section 2.
Ce modèle prend en
compte la notion de mot, distincte de la notion de token en raison notamment
des expressions multi-mots.

La plate-forme Outilex a mis au point deux formats nouveaux
pour la représentation du texte étiqueté. Le premier est le format binaire de
sortie de l'outil de consultation des lexiques (3.2). Il permet un
traitement rapide de textes de grande taille. Le deuxième,
fsa.xml, est destiné à
l'échange de données (fig.~4). C'est une traduction en XML du
format fst2 d'Unitex.
\begin{figure}
\scriptsize
\hspace{1in}
\begin{minipage}[t]{2.25in}
\tiny
\begin{verbatim}
  <q id="31" pos="138">
   (...)
   <tr to="33">
    <lex>
     <form>procès</form>
     <lem>procès</lem>
     <pos v="noun"/>
     <f n="proper" v="false"/>
     <f n="gender" v="m"/>
    </lex>
   </tr>
\end{verbatim}
\end{minipage}
\ 
\begin{minipage}[t]{2.5in}
\tiny
\begin{verbatim}
   <tr to="35">
    <lex>
     <form>procès-verbaux</form>
     <lem>procès-verbal</lem>
     <pos v="noun"/>
     <f n="proper" v="false"/>
     <f n="gender" v="m"/>
     <f n="number" v="p"/>
    </lex>
   </tr>
  </q>
\end{verbatim}
\end{minipage}
\vspace{4mm}
  \centering
  \includegraphics[width=3cm,height=17cm,angle=-90]{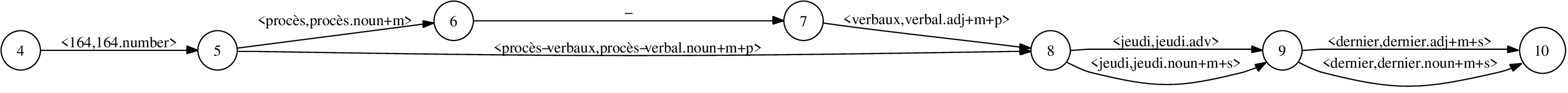}
\caption{Extrait d'un automate acyclique représentant un texte étiqueté}
\label{fig:fsa.xml}
\end{figure}
\begin{comment}
\begin{figure}
\footnotesize
\begin{verbatim}
  <q id="31" pos="138">
   (...)
   <tr to="33">
    <lex>
     <form>procès</form>
     <lem>procès</lem>
     <pos v="noun"/>
     <f n="proper" v="false"/>
     <f n="gender" v="m"/>
    </lex>

   </tr>
   <tr to="35">
    <lex>
     <form>procès-verbaux</form>
     <lem>procès-verbal</lem>
     <pos v="noun"/>
     <f n="proper" v="false"/>
     <f n="gender" v="m"/>
     <f n="number" v="p"/>
    </lex>
   </tr>
  </q>
\end{verbatim}
\caption{Extrait d'un automate acyclique représentant un texte étiqueté}
\end{figure}
\begin{figure}[h]
  \centering
  \includegraphics[width=3cm,height=17cm,angle=-90]{police.ps}
  \caption{Automate du texte}
\end{figure}
\end{comment}
Le projet MAF propose un format voisin.
Des fonctionnalités d'import/export entre ces deux formats sont prévues.
Un convertisseur entre le format binaire et le format fsa.xml est disponible.
De plus, les deux formats peuvent être exportés vers le format dot
\cite{gansner00open}.
\section{Traitement par grammaires}
Lors de l'étiquetage de mots par lexique, les étiquettes sont assignées aux mots
d'une façon indépendante du contexte. Cette procédure est généralement complétée
d'une façon ou d'une autre, dans les applications, par la prise en compte de
contraintes sur des séquences de mots, et donc de "grammaires", au sens de
ressources linguistiques spécifiant formellement de telles contraintes. Les
formalismes grammaticaux étant la tour de Babel du traitement des langues
naturelles, la plate-forme Outilex mise sur un formalisme minimal, qu'on peut
résumer en trois points :
\begin{itemize}
\item pour la représentation des mots et des paradigmes de mots, la notion de masque
lexical \cite{BlancDister:Recital2004}, spécification d'un ensemble de mots par un
ensemble de traits ;
\item pour la représentation des contraintes sur les séquences, la notion de réseau de
transitions récursif (RTN), outil purement formel, dépourvu de toute notion
linguistique, au même titre que les transducteurs finis ou les grammaires
algébriques ;
\item les automates constituant les RTN peuvent être des
transducteurs, c'est-à-dire comporter des sorties, utiles par exemple pour
insérer des balises dans les textes et formaliser ainsi des relations entre les
segments identifiés.
\end{itemize}
Ces points permettent de construire des grammaires locales au sens de
\cite{Gross:1993}, \cite{Gross:97:mit}.
Ce formalisme est utilisé
dans des situations variées : extraction d'informations
\cite{Poibeau:2001},\cite{Nakamura:2005},
reconnaissance d'entités nommées \cite{Krstev:LTC2005},
identification de structures grammaticales \cite{Mason:2004}, \cite{Danlos:IJCNLP2005}...
avec pour chacune de ces applications des taux de rappel et de précision qui
égalent l'état de l'art du domaine.
Nous avons ajouté la possibilité de pondérer les transitions, afin de permettre
la réalisation de systèmes hybrides utilisant à la fois des méthodes
statistiques et des méthodes fondées sur des ressources linguistiques. Nous
appelons le formalisme obtenu réseau de transitions récursif pondérées (WRTN).
\begin{figure}[h]
\centering
\includegraphics[width=10cm]{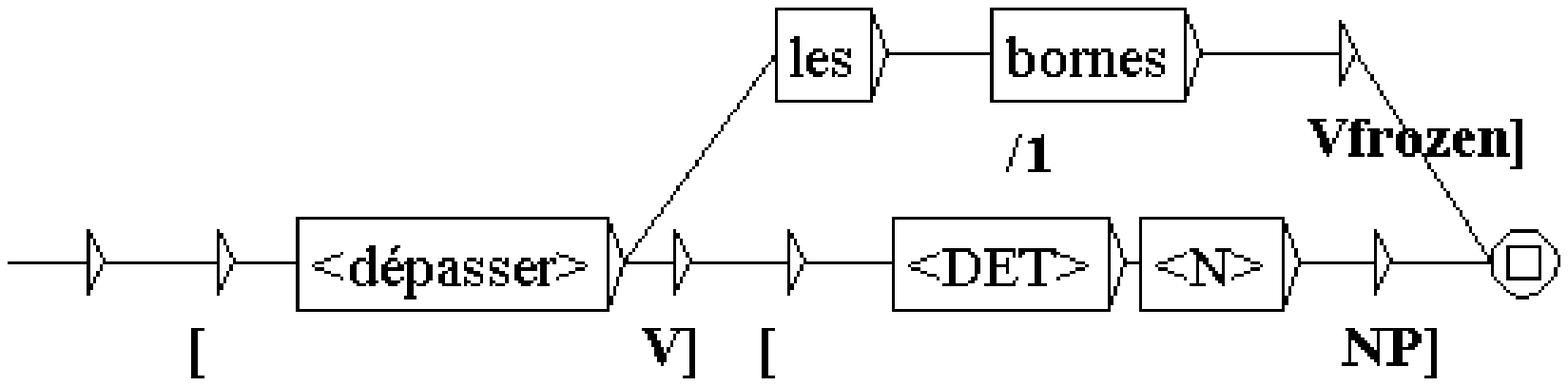}
\caption{Exemple de WRTN}
\end{figure}
Dans la fig.~5, par exemple, le chemin de poids 1, qui correspond à l'analyse
avec expression figée, est prioritaire par rapport à l'autre (de poids
0 par défaut) correspondant à l'analyse compositionnelle.
\subsection{Opérations sur les grammaires}
Les WRTN sont construits sous la forme de graphes à l'aide d'un éditeur
et sont sauvegardés dans un format XML appelé xgrf, élaboré à
partir de \cite{Sastre05}.
Ces grammaires sont ensuite compilées dans un format XML appelé wrtn, plus
adéquat aux traitements informatiques.
Au cours de cette opération, chaque graphe est optimisé par les
opérations d'émondation, suppression des transitions vides, 
déterminisation et minimisation.
Il est également possible de transformer une grammaire
en un transducteur fini équivalent, en faisant remonter les sous-graphes dans
le graphe principal, éventuellement à une approximation près. Le résultat occupe
plus d'espace mémoire mais accélère les traitements.
Outilex offre la possibilité de transcoder les graphes du format grf (Unitex)
au format xgrf (et inversement) et de les exporter vers le format dot \cite{gansner00open}.
\subsection{Traitements utilisant des grammaires}
Les traitements utilisant des grammaires identifient dans le texte étiqueté
les occurrences des motifs représentés dans les grammaires et peuvent produire
plusieurs types de résultat :
\begin{itemize}
  \item une concordance,
  \item une modification du texte linéaire ou de l'automate du texte,
  \item ou une forêt d'arbres d'analyse.
\end{itemize}
Tous ces traitements reposent sur un même moteur d'analyse, utilisant l'algorithme
d'Earley \cite{Earley70} adapté pour traiter d'une part des
WRTN (au lieu de grammaires algébriques) et d'autre part un texte sous forme d'automate acyclique
(au lieu d'une séquence de mots).
Notre analyseur fournit comme résultat une forêt partagée d'arbres d'analyse
pondérés pour chaque phrase analysée, les noeuds des arbres étant décorés
par les éventuelles sorties présentes dans la grammaire.
Appliqué au corpus de dépêches AFP avec une grammaire des groupes
nominaux inspirée de \cite{Paumier:2003}, il a traité 12 466 mots par seconde
et trouvé 39 468 occurrences.
\subsubsection{Présentation de concordances}
Nous avons développé un concordancier qui permet de lister dans leur contexte
d'apparition les différentes occurrences des motifs décrits par la grammaire. 
La taille des contextes gauche et droit peut être paramétrée par
l'utilisateur.
Les concordances peuvent être classées soit suivant leur ordre d'apparition
dans le texte, soit par ordre lexicographique.
\subsubsection{Application d'un transducteur au texte}
Nous avons également développé une fonctionnalité d'application d'un
transducteur sur le texte produisant un texte brut comportant les sorties
spécifiées dans la grammaire\footnote{Les sorties peuvent au choix être
insérées dans le texte d'origine ou remplacer les segments
reconnus.}.
Dans le cas de grammaires pondérées, les poids fournissent un critère de
filtrage entre plusieurs analyses concurrentes.
L'analyse retenue est celle dont le chemin a le poids le
plus élevé.  Un critère supplémentaire sur la longueur des séquences reconnues
peut également être utilisé.

Pour des traitements plus complexes, une variante de cette fonctionnalité
produit en sortie un automate correspondant à l'automate du texte auquel sont
rajoutées de nouvelles transitions étiquetées par les sorties de la grammaire.
Ce procédé, facilement itérable, permet de reconnaitre des segments de plus en
plus grands.
Il peut également être utilisé comme complément à l'étiquetage
morpho-syntaxique pour la reconnaissance d'unités
lexicales semi-figées dont les variations sont trop complexes pour être
énumérées sous forme de liste mais qui peuvent être décrites dans des
grammaires locales. Par exemple, la figure~6 présente
l'automate de la phrase précédente après l'application de la grammaire des
adverbes de temps de M. Gross.
\begin{figure}[h]
  \centering
  \includegraphics[width=3cm,height=17cm,angle=-90]{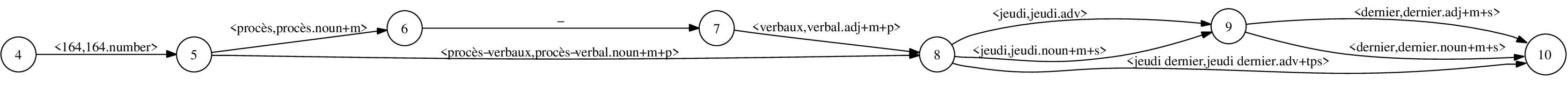}
  \caption{Résultat de l'application d'un transducteur sur l'automate du texte}
\end{figure}
\subsubsection{Grammaire d'unification pour l'analyse syntaxique}
Nous proposons enfin un module d'analyse syntaxique plus "fine" basé sur des
grammaires d'unification dans le formalisme des WRTN décorés
\cite{BlancConstant:2005}.  Ce formalisme allie au WRTN des équations
fonctionnelles sur les traits permettant ainsi de formaliser des phénomènes 
syntaxique d'extraction ou de coréférence.
Le résultat de notre analyseur consiste en une forêt partagée d'arbres syntaxiques ;
à chaque arbre est associée une structure de traits dans laquelle sont représentées
les relations grammaticales entre les constituants syntaxiques qui ont été identifiés
durant l'analyse.
La présentation de cet analyseur dépasse le cadre de cet article.
\section{Gestion de ressources linguistiques}
La réutilisation de
lexiques présuppose une certaine flexibilité\footnote{Un lexique n'est pas une ressource
statique. En raison de l'évolution de la langue,
et en particulier de la langue technique, des mises à jour
régulières sont nécessaires ; une nouvelle application d'un lexique peut mettre
en jeu la sélection d'un vocabulaire spécifique au domaine. Il en est de même
des grammaires, pour peu qu'elles soient lexicalisées.}.
La gestion des lexiques et grammaires, dont l'IGM et Systran sont
spécialistes, repose sur deux points :
construction et maintenance manuelles des ressources dans un
format lisible, sous la forme de petites unités cumulables (entrées lexicales,
par exemple) ; compilation en un format directement opérationnel. Cependant,
les manuels de
traitement des langues naturelles, même généralistes et réputés, comme
\cite{JurafskyMartin:2000}, ne traitent pas ces techniques, qui nécessitent une
collaboration étroite d'informaticiens et de linguistes ;
et peu de systèmes fournissent les
fonctionnalités requises (Xelda, Intex, Unitex).
La plate-forme Outilex propose donc un jeu complet d'outils de gestion de
ressources linguistiques.
\subsection{Flexion automatique des lexiques}
Le module de flexion automatique prend en entrée un lexique de lemmes et des
règles de flexion et produit en sortie un lexique
fléchi. Par exemple, le verbe \textit{carry} appartient à la classe de flexion de la
fig.~\ref{fig:flexion} et le module produit entre autres
la forme fléchie \textit{carries} avec le code "prs\_3s" (troisième personne
du singulier au présent).
\begin{figure}[h]
\begin{minipage}[t]{3in}
\tiny
\begin{verbatim}
<Paradigm code="4">
   <StemTrigger>y</StemTrigger>
   <Inflection caseId="inf"><Form>y</Form></Inflection>
   <Inflection caseId="prs_1s"><Form>y</Form></Inflection>
   <Inflection caseId="prs_2s"><Form>y</Form></Inflection>
   <Inflection caseId="prs_3s"><Form>ies</Form></Inflection>
\end{verbatim}
\end{minipage}
\ 
\begin{minipage}[t]{3in}
\tiny
\begin{verbatim}
   <Inflection caseId="prs_p"><Form>y</Form></Inflection>
   <Inflection caseId="prt"><Form>ied</Form></Inflection>
   <Inflection caseId="imp"><Form>y</Form></Inflection>
   <Inflection caseId="ppt"><Form>ied</Form></Inflection>
   <Inflection caseId="ppr"><Form>ying</Form></Inflection>
</Paradigm>
\end{verbatim}
\end{minipage}
\caption{Règle de flexion}
\label{fig:flexion}
\end{figure}
\subsection{Indexation des lexiques}
Afin d'accélérer leur consultation (cf. 3.2), les lexiques sont indexés sur les
formes fléchies en utilisant une représentation par automate fini minimal
\cite{Revuz:1991} qui permet de les comprimer tout en offrant un accès rapide à
l'information.  Le format binaire obtenu (fichier .idx) est adapté aux lexiques
à jeu d'étiquettes riche.
Le tableau suivant décrit
l'indexation du DELAF français \cite{Courtois:1990}
et la taille du fichier obtenu après l'indexation
du même lexique par les outils équivalents délivrés
avec Unitex\footnote{Les temps de calcul ne sont pas indiqués ici. Le programme
d'Unitex étant trop demandeur en ressources mémoire, nous avons dû lancer l'opération
sur une autre machine.}.

\begin{tabular}{|l|l|l|l|l|l|l|}
\hline
\# formes & \# formes  & taille      & taille XML & taille & temps        & taille\\
fléchies  & canoniques & DELA (utf8) & (xml.gz)   & idx    & d'indexation & Unitex\\ \hline
1264170   & 122035     & 35 Mo       & 220 Mo (5.3 Mo) & 9.5 Mo & 1m03s   & 59 Mo \\
\hline
\end{tabular}
\section{Conclusion et perspectives}
La plate-forme Outilex, dans sa version préliminaire actuelle, 
effectue toutes les opérations fondamentales du traitement automatique du
texte écrit : traitements sans lexiques, exploitation de lexiques et de
grammaires, gestion de ressources linguistiques. Les données manipulées sont
structurées dans des formats XML, et également dans d'autres formats plus
compacts, soit lisibles soit binaires ; les
convertisseurs de formats nécessaires sont inclus dans la plate-forme ; le
formalisme des WRTN permet de combiner des méthodes statistiques avec des
méthodes fondées sur des ressources linguistiques. Enfin, des lexiques issus
du LADL, construits manuellement et d'une couverture substantielle seront
distribuées avec la plate-forme sous licence LGPL-LR.

Le développement de la plate-forme a nécessité une expertise conjointe des
éléments informatiques et linguistiques du problème ; il a pris en compte les
besoins de la recherche fondamentale et ceux des applications. Nous pensons
qu'il n'aurait pas été possible sans un consortium aussi varié.
Au-delà de la fin du projet, l'avenir de la plate-forme Outilex est conçu dans
le cadre du développement collaboratif. Nous espérons que la plate-forme
actuelle, déjà compatible avec de nombreuses langues, sera étendue à d'autres et
enrichies en fonctionnalités nouvelles.
%%================================================================
%%================================================================
%% Note : si l'on préfère éviter de factoriser les crossrefs :
%% bibtex -min-crossrefs 99 taln-exemple
%%================================================================
\bibliographystyle{taln2002}
\bibliography{biblio}
%%================================================================
\end{document}